\documentclass[runningheads]{llncs}
\usepackage{graphicx}
\usepackage{amsmath}
\usepackage{amssymb} 
\usepackage{bm}
\usepackage{color}
\usepackage[width=122mm,left=12mm,paperwidth=146mm,height=193mm,top=12mm,paperheight=217mm]{geometry}
\usepackage{times}
\usepackage{epsfig}
\usepackage{url}
\usepackage{subfigure}
\usepackage{algorithm}
\usepackage{algorithmic}
\usepackage{enumerate}
\usepackage{multirow}
\usepackage{verbatim}
\usepackage{makecell}
\usepackage{microtype}

\usepackage[breaklinks=true,bookmarks=false]{hyperref}
\begin{document}
\pagestyle{headings}
\mainmatter
\title{Deep Deformation Network for Object Landmark Localization} 
\author{Xiang Yu, Feng Zhou and Manmohan Chandraker}
\institute{NEC Laboratories America, Department of Media Analytics\\
\email{\{xiangyu,manu\}@nec-labs.com, zhfe99@gmail.com}
}
\maketitle

\newcommand{\A}{\mathbf{A}}
\newcommand{\B}{\mathbf{B}}
\newcommand{\C}{\mathbf{C}}
\newcommand{\tC}{\tilde{\mathbf{C}}}
\newcommand{\D}{\mathbf{D}}
\newcommand{\E}{\mathbf{E}}
\newcommand{\F}{\mathbf{F}}
\newcommand{\G}{\mathbf{G}}
\newcommand{\bH}{\mathbf{H}}
\newcommand{\I}{\mathbf{I}}
\newcommand{\J}{\mathbf{J}}
\newcommand{\K}{\mathbf{K}}
\newcommand{\bL}{\mathbf{L}}
\newcommand{\M}{\mathbf{M}}
\newcommand{\bO}{\mathbf{O}}
\newcommand{\bP}{\mathbf{P}}
\newcommand{\Q}{\mathbf{Q}}
\newcommand{\R}{\mathbf{R}}
\newcommand{\bS}{\mathbf{S}}
\newcommand{\T}{\mathbf{T}}
\newcommand{\U}{\mathbf{U}}
\newcommand{\V}{\mathbf{V}}
\newcommand{\W}{\mathbf{W}}
\newcommand{\X}{\mathbf{X}}
\newcommand{\Y}{\mathbf{Y}}
\newcommand{\tY}{\tilde{\mathbf{Y}}}
\newcommand{\Z}{\mathbf{Z}}

\newcommand{\ba}{\mathbf{a}}
\newcommand{\bb}{\mathbf{b}}
\newcommand{\bc}{\mathbf{c}}
\newcommand{\bd}{\mathbf{d}}
\newcommand{\be}{\mathbf{e}}
\newcommand{\f}{\mathbf{f}}
\newcommand{\g}{\mathbf{g}}
\newcommand{\h}{\mathbf{h}}
\newcommand{\bk}{\mathbf{k}}
\newcommand{\m}{\mathbf{m}}
\newcommand{\n}{\mathbf{n}}
\newcommand{\p}{\mathbf{p}}
\newcommand{\q}{\mathbf{q}}
\newcommand{\br}{\mathbf{r}}
\newcommand{\s}{\mathbf{s}}
\newcommand{\bt}{\mathbf{t}}
\newcommand{\bu}{\mathbf{u}}
\newcommand{\bv}{\mathbf{v}}
\newcommand{\w}{\mathbf{w}}
\newcommand{\x}{\mathbf{x}}
\newcommand{\y}{\mathbf{y}}
\newcommand{\z}{\mathbf{z}}
\newcommand{\tz}{\tilde{\mathbf{z}}}
\newcommand{\ty}{\tilde{\mathbf{y}}}

\newcommand{\cA}{\mathcal{A}}
\newcommand{\cC}{\mathcal{C}}
\newcommand{\cD}{\mathcal{D}}
\newcommand{\cF}{\mathcal{F}}
\newcommand{\cG}{\mathcal{G}}
\newcommand{\cH}{\mathcal{H}}
\newcommand{\cI}{\mathcal{I}}
\newcommand{\cL}{\mathcal{L}}
\newcommand{\cM}{\mathcal{M}}
\newcommand{\cN}{\mathcal{N}}
\newcommand{\cP}{\mathcal{P}}
\newcommand{\cQ}{\mathcal{Q}}
\newcommand{\cR}{\mathcal{R}}
\newcommand{\cS}{\mathcal{S}}
\newcommand{\cT}{\mathcal{T}}
\newcommand{\cU}{\mathcal{U}}
\newcommand{\cV}{\mathcal{V}}
\newcommand{\cX}{\mathcal{X}}
\newcommand{\cY}{\mathcal{Y}}
\newcommand{\cZ}{\mathcal{Z}}

\newcommand{\bp}{\bar{\p}}
\newcommand{\BP}{\bar{\bP}}

\def\bphi{\mbox{\boldmath $\phi$}}
\def\bdelta{\mbox{\boldmath $\delta$}}
\def\blambda{\mbox{\boldmath $\lambda$}}
\def\bLambda{\mbox{\boldmath $\Lambda$}}
\def\bPi{\mbox{\boldmath $\Pi$}}
\newcommand{\bTheta}{\mathbf{\Theta}}
\newcommand{\bSigma}{\mathbf{\Sigma}}
\newcommand{\one}{\mathbf{1}}
\newcommand{\zero}{\mathbf{0}}
\newcommand{\epi}{\mathbf{epi}}
\newcommand{\dom}{\mathbf{dom}}
\newcommand{\real}{\mathbb{R}}
\newcommand{\bina}{\{0, 1\}}
\newcommand{\integer}{\mathbb{Z}}
\newcommand{\acc}{\text{acc}}
\newcommand{\err}{\text{err}}
\newcommand{\dist}{\text{dist}}
\newcommand{\convA}{\stackbin[\A]{}{*}}
\newcommand{\convAj}{\stackbin{\A_j}{*}}
\newcommand{\Searrow}{\rotatebox[origin=c]{-45}{$\Rightarrow$}}
\newcommand{\udots}{\mathinner{\mskip1mu\raise1pt\vbox{\kern7pt\hbox{.}}\mskip2mu\raise4pt\hbox{.}\mskip2mu\raise7pt\hbox{.}\mskip1mu}}

\def\astm{\ \dot{*}_m  }
\def\astmm{\ \ddot{*}_m  }
\def\astl{\ \dot{*}_l \ }
\def\astll{\ \ddot{*}_l \ }
\def\astx{ \dot{*}_{n_x}  }
\def\astxx{ \ddot{*}_{n_x}  }
\def\asts{ \ \dot{*}_{n_s} }
\def\astss{ \ \ddot{*}_{n_s} }
\def\astsss{ \ \dddot{*}_{n_s} }
\def\astz{ \ \dot{*}_{n_z}  }
\def\astzz{ \ \ddot{*}_{n_z}  }

\newcommand{\Hl}{\overleftarrow{\bH}}
\newcommand{\Hr}{\overrightarrow{\bH}}
\newcommand{\dk}{\dot{k}}
\newcommand{\dd}{\dot{d}}
\newcommand{\hQ}{\hat{\Q}}

\newcommand\etal{{et al.}}
\newcommand\eg{\emph{e.g.}}
\newcommand\ie{\emph{i.e.}}
\newcommand\wrt{\emph{w.r.t.}}
\newcommand\aka{\emph{a.k.a.}}
\newcommand\etc{\emph{etc}}

\begin{abstract}
We propose a novel cascaded framework, namely deep deformation network (DDN), for localizing landmarks in non-rigid objects. The hallmarks of DDN are its incorporation of geometric constraints within a convolutional neural network (CNN) framework, ease and efficiency of training, as well as generality of application. A novel shape basis network (SBN) forms the first stage of the cascade, whereby landmarks are initialized by combining the benefits of CNN features and a learned shape basis to reduce the complexity of the highly nonlinear pose manifold. In the second stage, a point transformer network (PTN) estimates local deformation parameterized as thin-plate spline transformation for a finer refinement. Our framework does not incorporate either handcrafted features or part connectivity, which enables an end-to-end shape prediction pipeline during both training and testing. In contrast to prior cascaded networks for landmark localization that learn a mapping from feature space to landmark locations, we demonstrate that the regularization induced through geometric priors in the DDN makes it easier to train, yet produces superior results. The efficacy and generality of the architecture is demonstrated through state-of-the-art performances on several benchmarks for multiple tasks such as facial landmark localization, human body pose estimation and bird part localization.


\end{abstract}

\section{Introduction}
\label{sec:intro}

Consistent localization of semantically meaningful landmarks or keypoints in images forms a precursor to several important applications in computer vision, such as face recognition, human body pose estimation or 3D visualization. However, it remains a significant challenge due to the need to handle non-rigid shape deformations, appearance variations and occlusions. For instance, facial landmark localization must handle not only coarse variations such as head pose and illumination, but also finer ones such as expressions and skin tones. Human body pose estimation introduces additional challenges in the form of large layout changes of parts due to articulations. Objects such as birds display tremendous variations in both appearance across species, as well as shape within the same species (for example, a perched bird as opposed to a flying one), which renders accurate part localization a largely open problem.

Consequently, there have been a wide range of approaches to solve the problem, starting with those related to PCA-based shape constraints (such as active shape models \cite{cootes1995}, active appearance models \cite{cootes1998} and constrained local models \cite{Cristinacce2007}) and pictorial structures (such as DPM \cite{felzenszwalb2010,yang2011} and poselets \cite{bourdev2010}). In recent years, the advent of convolutional neural networks (CNNs) has led to significant gains in feature representation \cite{simonyan2014}. In particular, cascaded regression networks specifically designed for problems such as facial landmark localization \cite{sun2013,zhang2014,cfan2014} or human body pose estimation \cite{toshev2014} have led to improvements by exploiting problem structure at coarse and fine levels. But challenges for such frameworks have been the need for careful design and initialization, the difficulty of training complex cascades as well as the absence of learned geometric relationships.

\begin{figure}[t]
  \centering
  \includegraphics[width=0.98\textwidth]{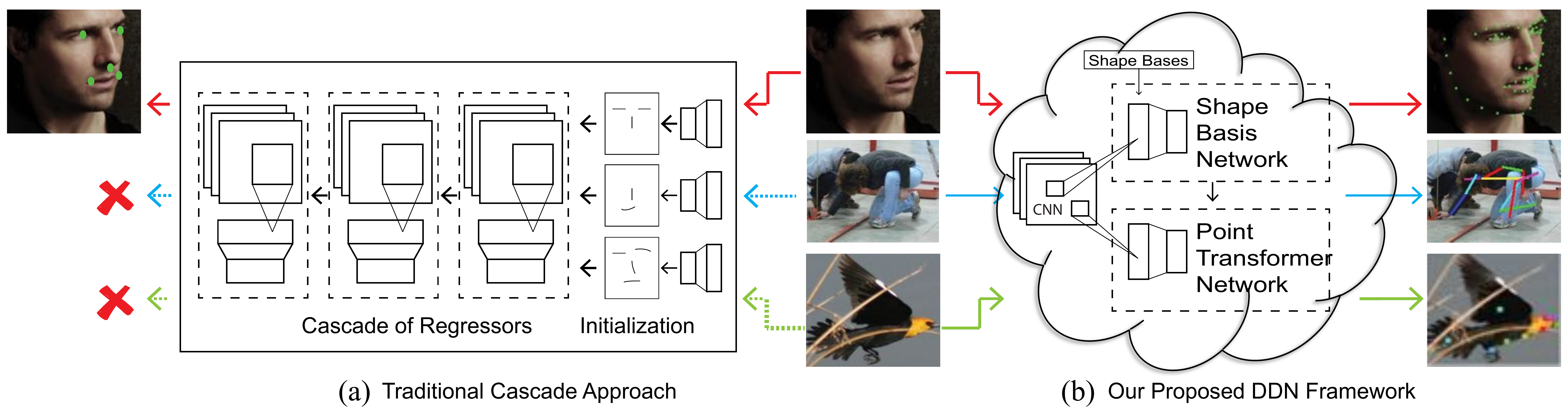}
  \caption{(a) Traditional CNN cascades use specialized initializations and directly map the features to landmark locations. This leads to prohibitively expensive training and testing times and frameworks that require complex design of cascade stages for different object types. (b) Our proposed Deep Deformation Network incorporates geometric constraints within the CNN framework. A shape basis network produces rapid global initializations and a point transformer network refines with local  non-rigid deformations. The entire framework is trainable end-to-end and results in state-of-the-art accuracy for several object types, while retaining the same network structure.}
  \label{fig:teaser}
\end{figure}

In this paper, we propose a novel cascaded framework, termed deep deformation network (DDN), that also decomposes landmark localization into coarse and fine localization stages. But in contrast to prior works, we do not train cascaded regressors to learn a mapping between CNN features and landmark locations. Rather, stages in our cascade explicitly account for the geometric structure of the problem within the CNN framework. We postulate that this has three advantages. First, our framework is easier to train and test, in contrast to previous cascaded regressors where proper initialization is required and necessitates training a battery of individual CNNs as sub-problems. Second, incorporation of geometric structures at both the coarse and fine levels regularizes the learning for each stage of the cascade by acting as spatial priors. Third, our cascade structure is general and still results in higher accuracy by learning part geometries and avoiding hard-coded connections of parts. These advantages are illustrated in Fig.~\ref{fig:teaser}.

Specifically, in Section \ref{sec:method}, we propose two distinct mechanisms to inject geometric knowledge into the problem of landmark localization. First, in Section \ref{sec:sbn}, we propose a shape basis network (SBN) to predict the optimal shape that lies on a low-rank manifold defined by the training samples. Our hypothesis is that shapes or landmarks for each object type reside close to a shape space, for which a low-rank decomposition reduces representation complexity and acts as regularization for learning. Note that unlike DPM, we do not define geometric connections among parts prior to localization, rather these relationships are learned. Further, even cascaded CNN frameworks such as \cite{sun2013} train individual CNNs for pre-defined relative localization of groups of parts within the first level of the cascade, which serves as initialization for later stages. Our SBN avoids such delicate considerations in favor of a learned basis that provides good global initializations. Second, in Section \ref{sec:ptn}, we propose a point transformer network (PTN) that learns the optimal local deformation in the form of a thin-plate spline (TPS) transformation that maps the initialized landmark to its final position.

A notable feature of our framework is its generality. Prior works explicitly design network structures to handle shape and appearance properties specific to object types such as faces, human bodies or birds. In contrast, our insights are quite general - a shape basis representation is suitable for regularized learning of a global initialization in a CNN framework, following which local deformations through learned TPS transformations can finely localize landmarks. We demonstrate this generality in Section \ref{sec:experiments} with extensive experiments on landmark localization that achieve state-of-the-art in accuracy for three distinct object types, namely faces, human bodies and birds, on several different benchmarks. We use the same CNN architectures for each of these experiments, with identical and novel but straightforward training mechanisms.

To summarize, the main contributions of this paper are:
\begin{itemize}
\item A novel cascaded CNN framework, called deep deformation network, for highly accurate landmark localization.
\item A shape basis network that learns a low-rank representation for global object shape.
\item A point transformer network that learns local non-rigid transformations for fine deformations, using the SBN output as an initialization.
\item A demonstration of the ease and generality of deep deformation network through state-of-the-art performance for several object types.
\end{itemize}

\section{Related Work}
\label{sec:related}

{\bf{Facial landmark localization}}
Facial landmark localization or face alignment is well-studied in computer vision. Models that impose a PCA shape basis have been proposed as Active Shape Models \cite{cootes1995} and its variants that account for holistic appearance \cite{cootes1998} and local patch appearance \cite{Cristinacce2007}. The non-convexity of the problem has been addressed through better optimization strategies that improve modeling for either the shape \cite{saragih2011,yu2013,pedersoli2014,yu2015} or appearance \cite{matthews2004,tzimiropoulos2013,cheng2013}. Exemplar consensus~\cite{belhumeur2011,yu2014} and graph matching~\cite{zhoufeng2013} show high accuracy on localization. Regression based methods~\cite{cao2012,dantone2012,xiong2013} that directly learn a mapping from the feature space to predict the landmark coordinates have been shown to perform better. Traditional regression-based methods rely on hand-craft features, for example, shape indexed feature \cite{cao2012}, SIFT \cite{xiong2013} or local binary feature \cite{ren2014}. Subsequent works such as \cite{vahid2014,lee2015} have also improved efficiency. The recent success of deep networks has inspired cascaded CNNs to jointly optimize over several facial parts \cite{sun2013}. Variants based on coarse-to-fine auto encoders \cite{cfan2014,zhu2015} and multi-task deep learning \cite{zhang2014,yangheng2015} have been proposed to further improve performance. In contrast, cascade stages in our deep deformation networks do not require careful design or initialization and explicitly account for both coarse and fine geometric transformations.

{\bf{Human body pose estimation}}
Estimating human pose is more challenging due to greater articulations. Pictorial structures is one of the early influential models for representing human body structure \cite{felzenszwalb2005}. The deformable part model (DPM) achieved significant progress in human body detection by combining pictorial structures with strong template features and latent-SVM learning \cite{felzenszwalb2010}. Yang and Ramanan extend the model by incorporating body part patterns \cite{yang2011}, while Wang and Li propose a tree-structured learning framework to achieve better performance against handcrafted part connections \cite{fwang2013}. Pischulin et al.~\cite{pishchulin2013} apply poselets \cite{bourdev2010} to generate mid-level features regularizing pictorial structures. Chen and Yuille \cite{chen2014} propose dependent pairwise relations with a graphical model for articulated pose estimation. Deep neural network based methods have resulted in better performances in this domain too. Toshev and Szegedy~\cite{toshev2014} propose cascaded CNN regressors, Tompson \etal~\cite{tompson2014} propose joint training for a CNN and a graphical model, while Fan \etal~\cite{fan2015} propose a dual-source deep network to combine local appearance with a holistic view. In contrast, our DDN also effectively learns part relationships while being easier to train and more efficient to evaluate.

{\bf Bird part localization}
Birds display significant appearance variations between classes and shape variations within the same class. An early work that incorporates a probabilistic model and user responses to localize bird parts is presented in \cite{wah2011}. Chai \etal~\cite{chai2013} apply symbiotic segmentation for part detection. The exemplar-based model of \cite{liu2013}, similar to \cite{belhumeur2011}, enforces pose and subcategory consistency to localize bird parts. Recently, CNN-based methods, for example, part-based R-CNN \cite{ning2014} and Deep LAC \cite{lin2015} have demonstrated significant performance improvements.

{\bf General pose estimation}
While the above works focus on a specific object domain, a few methods have been proposed towards pose estimation for general object categories. As a general framework, DPM has also been shown to be effective beyond human bodies for facial landmark localization \cite{zhu2012}. A successful example of more general pose estimation is the regression-based framework of \cite{dollar2010} and its variants such as \cite{artizzu2013,yan2014}. However, such methods are sensitive to initialization, which our framework avoids through an effective shape basis network. 

{\bf Learning transformations with CNNs}
Agrawal \etal~use a Siamese network to predict discretized rigid ego-motion transformations formulated as a classification problem \cite{agrawal2015}. Razavian \etal \cite{raz2015} analyzes generating spatial information with CNNs, of which our SBN and PTN are specific examples in designing the spatial constraints. Our point transformer network is inspired by the the spatial transformer network of \cite{jaderberg2015}. Similar to WarpNet \cite{angjoo2016}, we move beyond the motivation of spatial transformer as an attention mechanism driven by the classification objective, to predict a non-rigid transformation for geometric alignment. In contrast to WarpNet, we exploit both supervised and synthesized landmarks and use the point transformer network only for finer local deformations, while using the earlier stage of the cascade (the shape basis network) for global alignment.

\section{Proposed Method}
\label{sec:method}

\begin{figure}[t]
  \centering
  \includegraphics[width=0.96\textwidth]{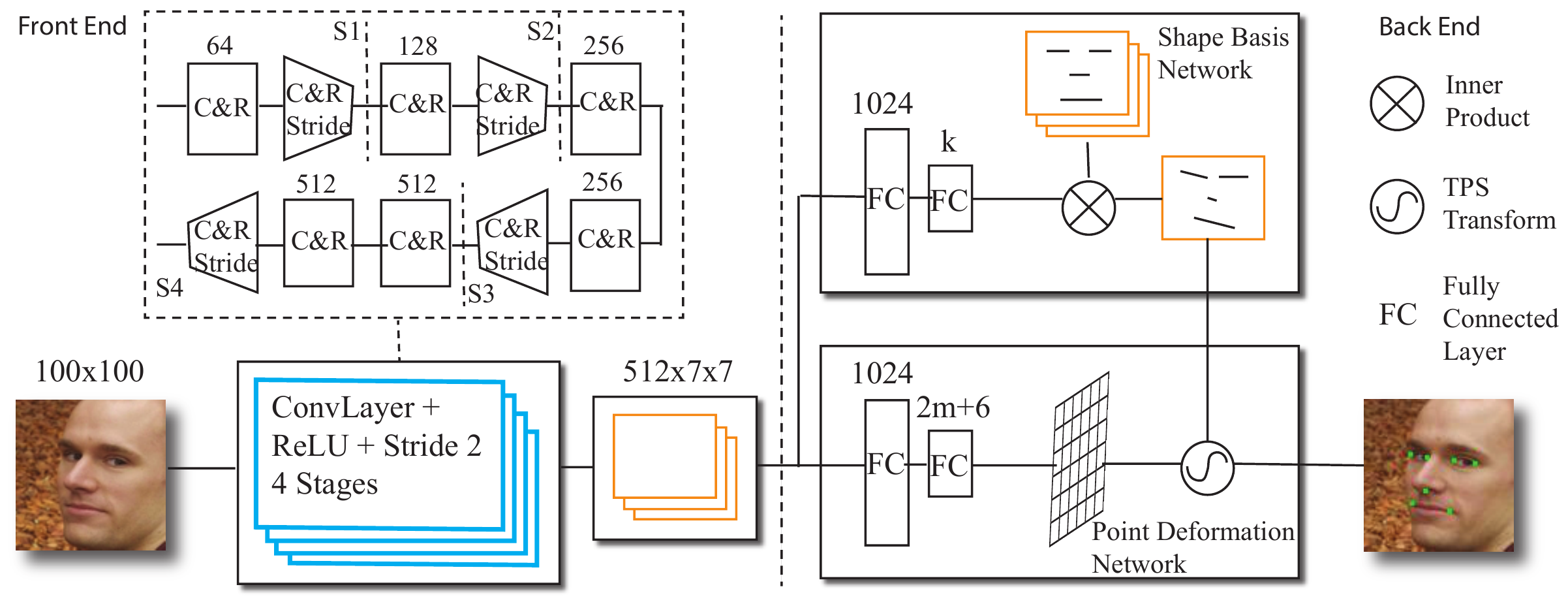}
  \caption{Overview of the proposed network architecture. Convolutional features are extracted with a 4-stage truncated VGG-16 network~\cite{simonyan2014}, denoted as S1 to S4 (C\&R stands for convolution and ReLU layers, with stride 2). The $512\times 7\times 7$ convolutional maps are sent to the Shape Basis Network and Point Transformer Network as input, where the PTN takes the SBN's output as another input to predict the final landmark positions.}
  \label{fig:flowchart}
\end{figure}

In this section, we present a general deep network to efficiently and accurately localize object landmarks. As shown in Fig.~\ref{fig:flowchart}, the network is composed of three components: 
\begin{itemize}
\item A modified VGGNet \cite{simonyan2014} to extract discriminative  features.
\item A Shape Basis Network (SBN) that combines a set of shape bases using weights generated from convolutional features to approximately localize the landmark.
\item A Point Transformer Network (PTN) for local refinement using a TPS transformation.
\end{itemize}
The entire network is trained end-to-end. We now introduce each of the above components in detail.

\subsection{Convolutional Feature Extraction}
\label{sec:feature}

For feature extraction, we adopt the VGG-16 network \cite{simonyan2014} that achieves state-of-the-art performance in various applications \cite{long2015,ren2015}. The upper left corner in Fig.~\ref{fig:flowchart} shows the network structure, where each stage consists of a convolutional layer followed by a ReLU unit. We apply 2 stride across the network. Similar to most localization algorithms, our network takes as input a region of interest cropped by an object detector. We scale input images to $100 \times 100$ resolution for facial landmark localization and $200 \times 200$ for body and bird pose estimation. Compared to classification and detection tasks, landmark localization requires extracting much finer or low-level image information. Therefore, we remove the last stage from the original $5$-stage VGG-16 network of \cite{simonyan2014}. We also experimented with using just the first three stages, but it performs worse than using four stages. In addition, we remove the pooling layers since we find they harm the localization performance. We hypothesize that shift-invariance achieved by pooling layers is helpful for recognition tasks, but it is beneficial to keep the learned features shift-sensitive for keypoint localization. Given an image at $100 \times 100$ resolution, the four-stage convolutional layers generate a $7 \times 7$ response map with $512$ output channels.

\subsection{Shape Basis Network}
\label{sec:sbn}
Let $\x \in \real^{d}$ denote the features extracted by the convolutional layers. Each training image is annotated with up to $n$ 2D landmarks, denoted $\y = [{\y^1}^\top, \cdots, {\y^n}^\top]^\top \in \real^{2n}$. To predict landmark locations, previous works such as \cite{sun2013,zhang2014} learn a direct mapping between CNN features $\x$ and ground-truth landmarks $\y$. Despite the success of these approaches, learning a vanilla regressor has limitations alluded to in Sec.~\ref{sec:intro}. First, a single linear model is not powerful enough to handle large shape variations. Although cascade regression can largely improve the performance, a proper initialization is required, which is non-trivial. Second, with limited data, training a large-capacity network without regularization from geometric constraints entails a high risk of overfitting.

Both of the above limitations are effectively addressed by our Shape Basis Network (SBN), which predicts optimal object shape that lies in a low-dimensional manifold defined by the training samples. Intuitively, while CNNs allow learning highly descriptive feature representations, frameworks such as active shape models \cite{cootes1995} have been historically effective in learning with non-linear pose manifolds. Our SBN provides an end-to-end trainable framework to combine these complementary advantages. Besides, for more challenging tasks, \ie human body, the highly articulated structure cannot be easily represented by multi-scale autoencoder or detection~\cite{cfan2014,zhu2015}. The SBN complete basis representation alleviates the problem, while retaining accuracy and low cost.
\begin{figure}[t]
  \centering
  \includegraphics[width=0.96\textwidth]{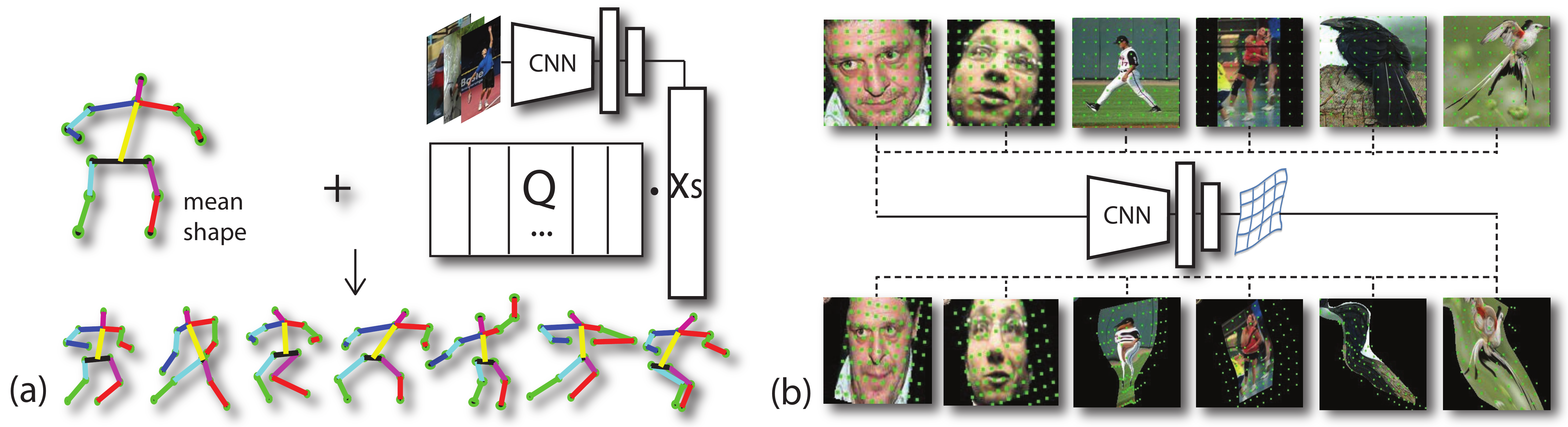}
  \caption{The workflow illustration of (a) Shape Basis Network and (b) Point Transformer Network. The SBN uses the mean shape and a PCA basis, along with the CNN features, to generate basis weights that compute the initialization for the PTN. The PTN uses the CNN features along with the output of the SBN to estimate non-rigid local deformations parameterized by TPS transformations.}
  \label{fig:sbnptn}
\end{figure}

More specifically, SBN predicts the shape $\y_s = [{\y_s^1}^\top, \cdots, {\y_s^n}^\top]^\top \in \real^{2n}$ as,
\begin{align}
\y_s = \bar{\y} + \mathbf{Q} \x_s \, , {\textrm{ where }} \x_s = f(\w_s, \x).
\label{eqn:pca}
\end{align}
Here $\bar{\y} \in \real^{2n}$ is the mean shape among all training inputs, the columns of $\Q \in \real^{2n \times k}$ store the top-$k$ orthogonal PCA bases and $f(\cdot)$ is a non-linear mapping parameterized by $\w_s$ that takes the CNN feature $\x$ as input to generate the basis weights $\x_s \in \real^k$ as output. We choose $k$ to preserve $99\%$ of the energy in the covariance matrix of the training inputs, $\sum_{\y} (\y - \bar{\y})(\y - \bar{\y})^\top$. As shown in upper-right corner of Fig.~\ref{fig:sbnptn}a, the mapping $f(\w_s, \x)$ is represented by stacking two fully connected layers, where the first layer encodes each input as a $1024$-D vector, which is further reduced to $k$ dimension by the second one. 
Conceptually, we jointly train SBN with other network components in an end-to-end manner. During backward propagation, given the gradient $\nabla \y_s \in \real^{2n}$, the gradient of $\x_s$ is available as $\Q^\top \nabla \y_s \in \real^{k}$. We then propagate this gradient back to update the parameters $\w_s$ for the fully connected layers as well as the lower convolutional layers.

In practice, we find it advantageous to pre-train the SBN on a simpler task and fine-tune with the later stages of the cascade. This is a shared motivation with curriculum learning \cite{BengioLCW09} and avoids the difficulties of training the whole network from scratch. Given the PCA shape model ($\bar{\y}$ and $\Q$) and the set of training images $\x$, we pre-train SBN to seek the optimal embedding $\x_s = f(\w_s, \x)$ such that the Euclidean distance between the prediction and the ground-truth ($\y$) is minimized\footnote{Strictly speaking, the loss is defined over a mini-batch of images.}, that is,
\begin{align}
\min_{\w_s} \quad \mathcal{F} = \|\y - (\bar{\y} + \mathbf{Q} \x_s) \|_{2}^{2} + \lambda \|\x_s\|_{2}^{2},
\label{eqn:sbn}
\end{align}
where $\lambda$ is a regularization factor that penalizes coefficients with large $l_{2}$ norm. We set $\lambda = 0.1$ in all experiments. To solve \eqref{eqn:sbn}, we propagate back the gradient over $\x_s$,
\begin{align}
\nabla_{\x_s} {\mathcal{F}} = 2\lambda \x_s - 2\mathbf{Q}^\top(\y - (\bar{\y}+\mathbf{Q}\x_s)),
\label{eqn:sbnerror}
\end{align}
to update the parameters $\w_s$ of the fully connected layers and the lower layers.

Thus, the SBN brings to bear the powerful CNN framework to generate embedding coefficients $\x_s$ that span the highly nonlinear pose manifold. The low-rank truncation inherent in the representation makes SBN clearly insufficient for localization on its own. Optimizing both the coefficients and PCA basis is possible. The orthogonality of the basis should be preserved, which causes extra efforts in tuning the network. As we assume, the role of SBN is to alleviate the difficulty of learning an embedding with large non-linear distortions and to reduce the complexity of shape deformations to be considered by the next stage of the cascade. 

\subsection{Point Transformer Network}
\label{sec:ptn}

Given the input feature $\x$, SBN generates the object landmark $\y_s$ as a linear combination of pre-defined shape bases. As discussed before, this prediction is limited by its assumption of a linear regression model. To handle more challenging pose variations, this section proposes the Point Transformer Network (PTN) to deform the initialized shape ($\y_s$) using a thin-plate spline (TPS) transformation to best match with ground-truth ($\y$). The refinement is not ideally local as global deformation also exists. Some neural methods~\cite{baltrusaitis2013} emphasize more on the local response map, while PTN incorporates global transformation into the overall deformation procedure. 

A TPS transformation consists of an affine transformation $\D \in \real^{2 \times 3}$ and a non-linear transformation parametrized by $m$ control points $\C = [\bc_1, \cdots, \bc_m] \in \real^{2 \times m}$ with the corresponding coefficients $\U = [\bu_1, \cdots, \bu_m] \in \real^{2 \times m}$ \cite{Bookstein_1989}. 
In our experiments, the control points form a $10 \times 10$ grid (that is, $m=100$). 
The TPS transformation for any 2D point $\z \in \real^2$ is defined as:
\begin{align}
g(\{\D, \U\}, \z) = \D \tz  + \sum_{j=1}^m \bu_{j} \phi(\|\z - \bc_j \|_{2}),
\label{eqn:tps}
\end{align}
where $\tz = \left[\z^\top, 1\right]^\top \in \real^{3}$ denotes $\z$ in homogeneous form and $\phi(d) = d^2 \log d$ is the radial basis function (RBF). 

Given convolutional features $\x$ and landmarks $\Y_s = [\y_s^1, \cdots, \y_s^n] \in \real^{2 \times n}$ initialized by the SBN of Sec.~\ref{sec:sbn}, the PTN seeks the optimal non-linear TPS mapping $f_p(\w_p, \x) = \{\D, \U\}$ to match the ground-truth $\Y = [\y^1, \cdots, \y^n] \in \real^{2 \times n}$. Similar to SBN, this mapping $f_p(\w_p, \x)$ is achieved by concatenating two fully connected layers, which first generate a $1024$-D intermediate representation, which is then used to compute $\w_p$. See Fig.~\ref{fig:sbnptn}b for an overview of PTN. Following \cite{Bookstein_1989}, PTN optimizes:
\begin{align}
\min_{\w_p} \ \sum_{i=1}^n \|\y^i - g(f_p(\w_p, \x), \y_s^i) \|_{2}^{2}+\gamma \int \|\nabla^{2} g\|_{2}^{2} d \y_s^i,
\label{eqn:tpsenergy}
\end{align}
where $\nabla^{2} g$ is the second-order derivative of transformation $g$ with respect to $\y_s^i$. The weight $\gamma$ is a trade-off between the transformation error and the bending energy. Substituting \eqref{eqn:tps} into the \eqref{eqn:tpsenergy} yields an equivalent objective,
\begin{align}
\min_{\w_p} \ \mathcal{E} = \|\Y - \D \tY_s - \U \mathbf{\Phi} \|_{F}^{2}+\gamma \text{tr}(\U \mathbf{\Phi} \mathbf{\Phi}^\top \U^\top),
\label{eqn:tpsobj}
\end{align}
where each element of the RBF kernel $\mathbf{\Phi} \in \real^{m \times n}$ computes $\phi_{j, i} = \phi( \|\y_s^i - \bc_j \| )$.

It is known that \eqref{eqn:tpsobj} can be optimized over the TPS parameters $\D$ and $\U$ in closed form for a pair of shapes. But in our case, these two parameters are generated by the non-linear mapping $f_p(\w_p, \x)$ from image features $\x$ on-the-fly. Thus, instead of computing the optimal solution, we optimize \eqref{eqn:tpsobj} over $\w_p$ using stochastic gradient descent. 

\begin{figure}[t]
  \centering
  \includegraphics[width=0.94\textwidth]{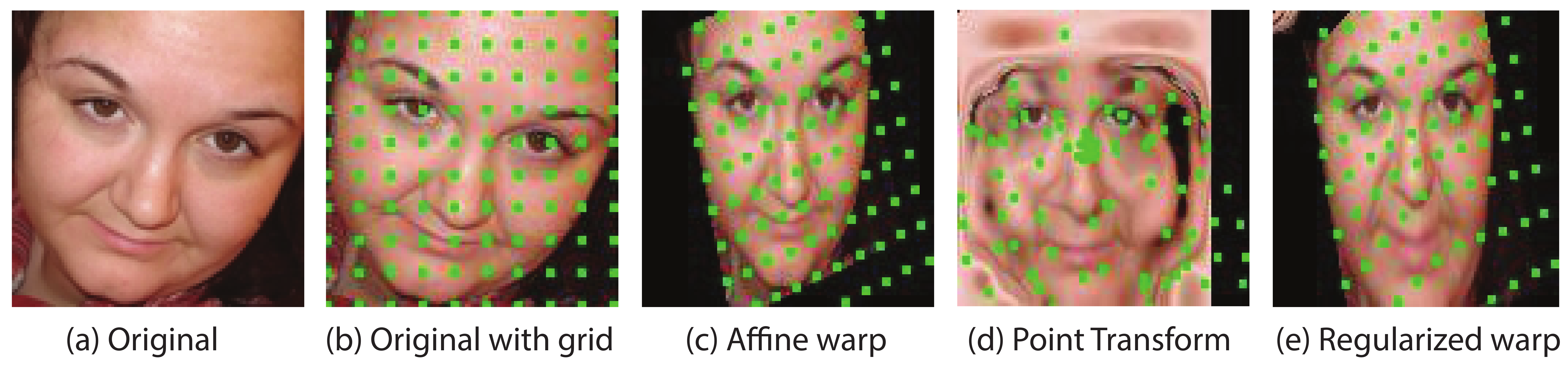}
  \caption{Illustration of different transformation models. (a) The input facial image. (b) The input image with uniform grid. (c) The solution of optimizing \eqref{eqn:tpsobj} by replacing $g(\cdot, \cdot)$ with an affine transformation. (d) The solution of point transformation by minimizing \eqref{eqn:tpsobj} without regularization. (e) The solution of regularized point transformation based on \eqref{eqn:regularize}.}
  \label{fig:warp}
\end{figure}

In practice, a key difficulty in training PTN stems from over-fitting of the non-linear mapping $f_p(\w_p, \x)$, since the number of parameters in $\w_p$ is larger than the number of labeled point pairs in each mini-batch. For instance, Fig.~\ref{fig:warp} visualizes the effect of using different spatial transformation to match an example with the mean face. By replacing the TPS in \eqref{eqn:tpsobj} with an affine transformation, PTN may warp the face to a frontal pose only to a limited extent (Fig.~\ref{fig:warp}c) due to out-of-plane head pose variations. Optimizing \eqref{eqn:tpsobj} with $n=68$ landmarks, PTN is able to more accurately align the landmarks with the mean face. As shown in Fig.~\ref{fig:warp}d, however, the estimated transformation is highly non-linear, which suggests a severe overfitting over the parameters of \eqref{eqn:tpsobj}.

To address overfitting, a common solution is to increase the regularization weight $\gamma$. However, a large $\gamma$ reduces the flexibility of the TPS transformation, which runs counter to PTN's purpose of generating highly non-rigid local deformations. Thus, we propose a control point grid regularization method to further constrain the point transformation. For each training image, we estimate the optimal TPS transformation $\{\bar{\D}, \bar{\U}\}$  from the mean shape $\bar{\y}$ to the ground truth $\y$ offline. Then this TPS transformation is applied on the $m$ control points $\bc_i$ to obtain their transformed locations $\y^i_{c} = g(\{\bar{\D}, \bar{\U} \}, \bc_i)$. We now obtain $m$ additional synthesized landmarks $\Y_c = [\y^1_c, \cdots, \y^m_c] \in \real^{2 \times m}$ with their original positions $\C = [\bc_1, \cdots, \bc_m] \in \real^{2 \times m}$. Finally, we define an improved loss incorporating the synthesized control points:
\begin{align}
 \mathcal{E}_c = \mathcal{E} + \varphi \| \Y_c - \D_c \tC  - \U_c \mathbf{\Phi}_c \|_{F}^{2}+ \psi \text{tr}(\U_c \mathbf{\Phi}_c \mathbf{\Phi}_c^\top \U_c^\top),
\label{eqn:regularize}
\end{align}
where the terms $\tC$, $\D_c$, $\U_c$ and $\mathbf{\Phi}_c$ are defined in a similar way as \eqref{eqn:tpsobj}. As shown in Fig.~\ref{fig:warp}e, the new loss $\mathcal{E}_c$ incorporates information from $m$ additional points, which helps to reduce overfitting and produces more stable TPS warps. In our experiments, the values for the involved parameters are $\gamma = 1$ and $\varphi = \psi = 0.4$.

To summarize, PTN forms the refinement stage for our cascade that generates highly non-rigid local deformations for a finer-level localization of landmarks. It is pertinent to note here that unlike the spatial transformer network \cite{jaderberg2015}, we directly optimize a geometric criterion rather than a classification objective. Similar to WarpNet \cite{angjoo2016}, we transform point sets rather than the entire image or dense feature maps, but go beyond it in striking a balance between supervised and synthesized landmark points. 

\section{Implementation Details}
\label{sec:implement}

We implement DDN using the Caffe platform \cite{jia2014caffe}. The three components of DDN shown in Fig.~\ref{fig:flowchart} -- the convolutional layers for extracting features $\x$, SBN for computing the intermediate landmarks $\y_s$ and PTN for generating the final position $\y_p$ -- can be trained from scratch end-to-end. However, for ease of training, we pre-train SBN and PTN separately, before a joint training. For each task of localizing landmarks on faces, human bodies and birds, we synthetically augment training data by randomly cropping, rotating and flipping images and landmarks. We use the standard hyper-parameters (0.9 for momentum and 0.004 for weight decay) in all experiments.

To pre-train SBN, we minimize \eqref{eqn:sbnerror} without the PTN part. For convolutional layers, we initialize with weights from the original VGG16 model. During the pre-train process, we first fix the convolutional weights and update the fully connected layers of SBN. When the error stops decreasing after $10$ epochs, the objective is relaxed to update both the convolutional layers and the fully connected layers. To pre-train PTN, we remove the SBN component from the network and replace the input $\y_s$ with the mean shape $\bar{\y}$. We fix the convolutional weights as the one pre-trained with SBN and train the fully connected layers in PTN only. After $10$ epochs, we train both the convolutional and fully connected layers together.

After pre-training, we combine SBN and PTN in a joint network, where the SBN provides shape input $\y_s$ to the PTN. The loss in \eqref{eqn:regularize} is evaluated at the end of PTN and is propagated back to update the fully connected and convolutional layers. During the joint training, we first update the weights of PTN by fixing the weights of SBN. Then the weights for SBN are relaxed and the entire network is updated jointly. 

\section{Experiments}
\label{sec:experiments}

We now evaluate DDN on its accuracy, efficiency and generality. Three landmark localization problems, faces, human bodies and birds are evaluated. The runtime is $1.3\pm 0.5$ ms for face landmark localization, $4.3\pm 3.6$ ms for human body part localization and $7.6\pm 2.4$ ms for bird part localization, on a Tesla K80 GPU with 12G memory, with an Intel 2.4GHz 8-core CPU machine.

{\bf Evaluation metrics}
For all the tasks, we use percentage of correctly localized keypoints (PCK)~\cite{yang2011} as the metric for evaluating localization accuracy. For the $j$-th sample in the test set of size $N$, PCK defines the predicted position of the $i$-th landmark, $\tilde{\y}_j^i$, to be correct if it falls within a threshold of the ground-truth position $\y_j^i$, that is, if
\begin{align}
  \|\y_{j}^i - \tilde{\y}_j^i \|_2 \leq \alpha \mathcal{D}, \label{eqn:pck}
\end{align}
where $\mathcal{D}$ is the reference normalizer, i.e. inter-ocular distance for face task, the maximum of the height and width of the bounding box for human body pose estimation and bird part localization, respectively. The parameter $\alpha$ controls the threshold for correctness. 

\subsection{Facial Landmark Localization}

To test face alignment in real scenarios, we choose the challenging 300 Faces in-the-Wild Challenge (300-W)~\cite{300w2013} as the main benchmark. It contains facial images with large head pose variations, expressions, types of background and occlusions. For instance, the first row of Fig.~\ref{fig:res} shows a few test examples from 300-W. The dataset is created from five well-known datasets -- LFPW~\cite{belhumeur2011}, AFW~\cite{zhu2012}, Helen~\cite{le2012}, XM2VTS~\cite{messer1999} and iBug~\cite{300w2013}. Training sets of Helen and LFPW are adopted as the overall training set. All the datasets use 68 landmark annotation. We report the relative normalized error on the entire 300-W database and further compare the PCK on the proposed component-wise methods.

\begin{table}[t]
\begin{center}
\setlength\tabcolsep{3.0pt}
\begin{tabular}{ccccccc}
\hline
\cline{1-7}
Dataset & ESR~\cite{cao2012} & SDM~\cite{xiong2013} & ERT~\cite{vahid2014} & LBF~\cite{ren2014} & cGPRT~\cite{lee2015} & DDN (Ours)\\
\hline
\cline{1-7}
300-W~\cite{300w2013} & 7.58 & 7.52 & 6.40 & 6.32 & 5.71 & \textbf{5.65}\\
\hline
\cline{1-7}
\end{tabular}
\end{center}
\caption{Comparison of accuracy on 300-W dataset.}
\label{tb:facenumber}
\end{table}

Table \ref{tb:facenumber} lists the accuracy of five state-of-the-art methods as reported in the corresponding literature-- explicit shape regression (ESR)~\cite{cao2012}, supervised descent method (SDM)~\cite{xiong2013}, ensemble of regression trees (ERT)~\cite{vahid2014}, regression of local binary features (LBF)~\cite{ren2014} and cascade of Gaussian process regression tree (cGPRT)~\cite{lee2015}. The benefit of the CNN feature representation allows our DDN framework to outperform all the other methods on 300-W. We note that the improvement over cGPRT is moderate. For face alignment, hand-crafted features such as SIFT feature are competitive with CNN features. This indicates that the extent of non-rigid warping in facial images, as opposed to human bodies or birds, is not significant enough to derive full advantage of the power of CNNs, which is also visualized in Fig.~\ref{fig:sbnptn}(b).

To provide further insights into DDN, we also evaluate the independent performances of the two components (SBN and PTN) in Table \ref{tb:facepck}. We note that SBN achieves poor localization compared to PTN. However, the PTN is limited to in-plane transformations. Thus, using SBN as an initialization, the combined DDN framework consistently outperforms the two independent networks. 

To illustrate the need for non-rigid TPS transformations, we modify the PTN to use an affine transformation. The network is denoted as a-DDN in Table ~\ref{tb:facepck}. The performance is worse than DDN, which indicates that the flexibility of representing non-rigid transformations is essential. Visual results in the first row of Fig.~\ref{fig:res} show that our method is robust to some degree of illumination changes, head pose variations and partial occlusions. The first failure case in the red box of the first row shows that distortions occur when large parts of the face are completely occluded. Note that in the second failure example, DDN actually adapts well to strong expression change, but is confused by a larger contrast for the teeth than the upper lip.

\begin{table}[t]
\begin{center}
\setlength\tabcolsep{6pt}
\begin{tabular}{ccccccccc}
    \hline
    \cline{1-9}
    Method & \multicolumn{2}{c}{Helen~\cite{le2012}} & \multicolumn{2}{c}{LFPW~\cite{belhumeur2011}} & \multicolumn{2}{c}{AFW~\cite{zhu2012}} & \multicolumn{2}{c}{iBug~\cite{300w2013}}\\
    \hline
    $\alpha$ & 0.05 & 0.10 & 0.05 & 0.10 & 0.05 & 0.10 & 0.05 & 0.10 \\
    \hline
    \cline{1-9}
    SBN (Ours)  & 51.4 & 87.0 & 48.1 & 84.0 & 36.4 & 74.0 & 22.6 & 57.3\\
    PTN (Ours) & 81.4 & 96.4 & 63.8 & 91.3 & 57.4 & 90.5 & 49.1 & 86.5\\
    a-DDN (Ours) & 67.8 & 93.5 & 55.5 & 89.3 & 44.3 & 82.9 & 38.7 & 79.3\\
    DDN (Ours) & \textbf{85.2} & \textbf{96.5} & \textbf{64.1} & \textbf{91.6} & \textbf{59.5} & \textbf{90.6} & \textbf{56.6} &  \textbf{88.9}\\
    \hline
    \cline{1-9}
\end{tabular}
\end{center}
\caption{Comparison of the PCK (\%) scores on different face datasets. Each component (SBN and PTN) is evaluated, a-DDN and DDN use different transformations (a-DDN for affine transformation and DDN for TPS).}
\label{tb:facepck}
\end{table}

\subsection{Human Body Pose Estimation}
Compared to facial landmarks, localization of human body parts is more challenging due to the greater degrees of freedom from the articulation of body joints. To evaluate our method, we use the Leeds Sports Pose (LSP) dataset~\cite{Johnson2010}, which has been widely used as the benchmark for human pose estimation. The original LSP dataset contains $2,000$ images of sportspersons gathered from Flickr, 1000 for training and 1000 for testing. Each image is annotated with $14$ joint locations, where left and right joints are consistently labelled from a person-centric viewpoint. To enhance the training procedure, we also use the extended LSP dataset~\cite{Johnson2011}, which contains 10,000 images labeled for training, which is the same setup used by most of the baselines in our comparisons.

We compare the proposed DDN with five state-of-the-art methods publicly reported on LSP. Two of them~\cite{fwang2013,pishchulin2013} are traditional DPM-based methods that model the appearance and shape of body joints in a tree-structure model. The other three~\cite{tompson2014,chen2014,fan2015} utilize CNNs for human pose estimation combined with certain canonical frameworks. Our method also uses convolutional features, but proposes two new network structures, SBN and PTN, which can be integrated for end-to-end training and testing. This is in contrast to  previous CNN-based methods that include graphical models or pictorial structures that introduce extra inference cost in both training and testing.

\begin{figure}[t]
\includegraphics[width=0.323\textwidth]{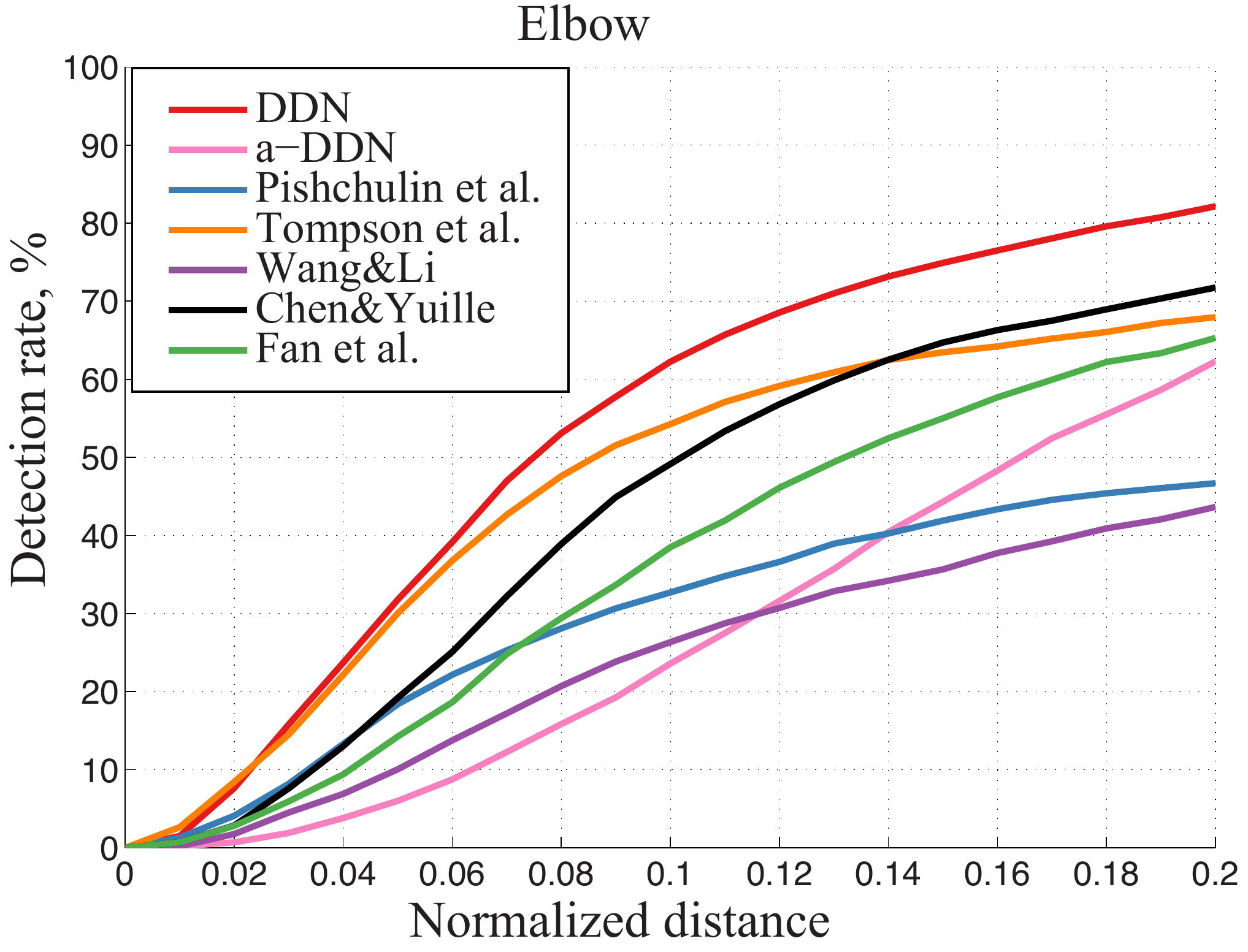}
\includegraphics[width=0.323\textwidth]{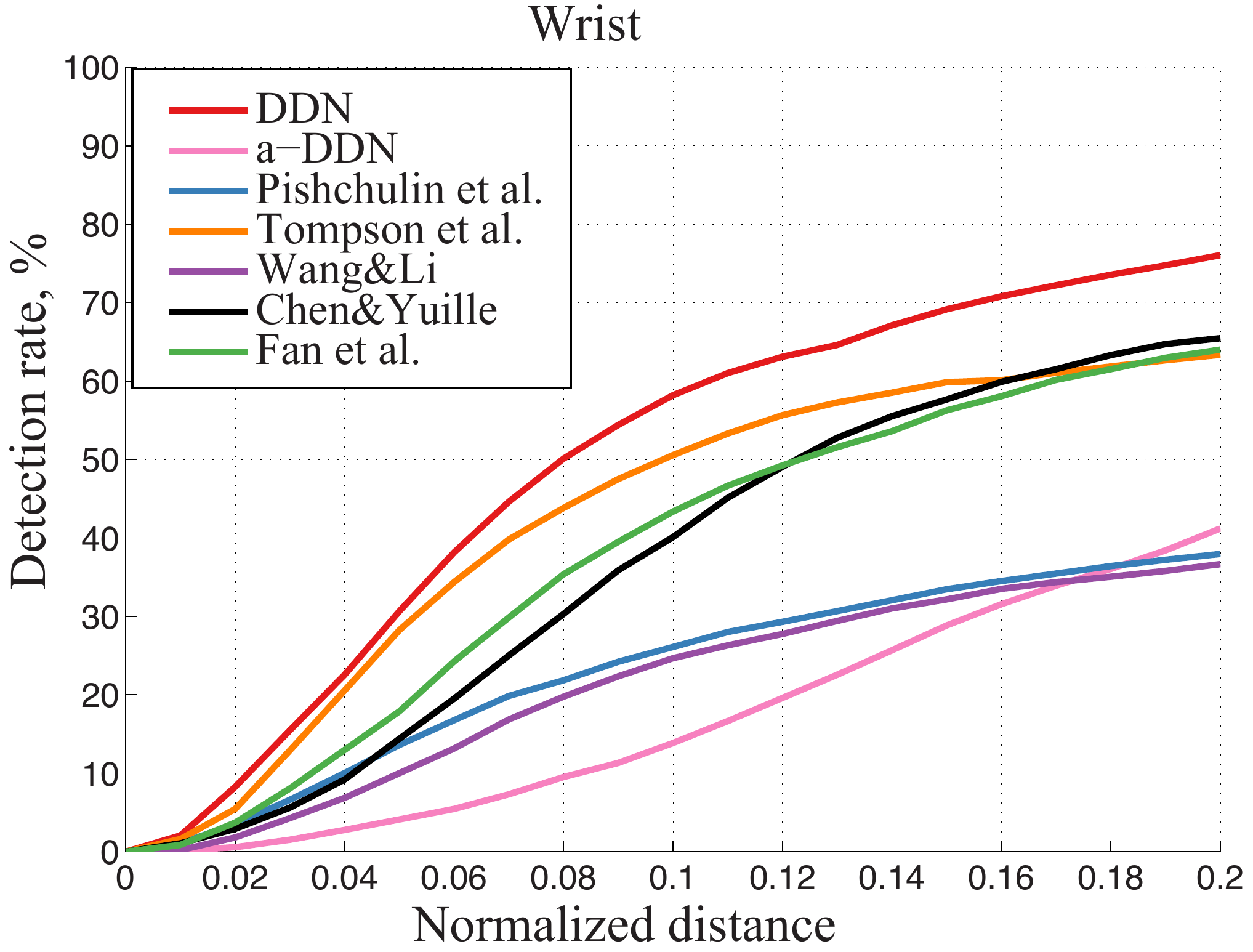}
\includegraphics[width=0.323\textwidth]{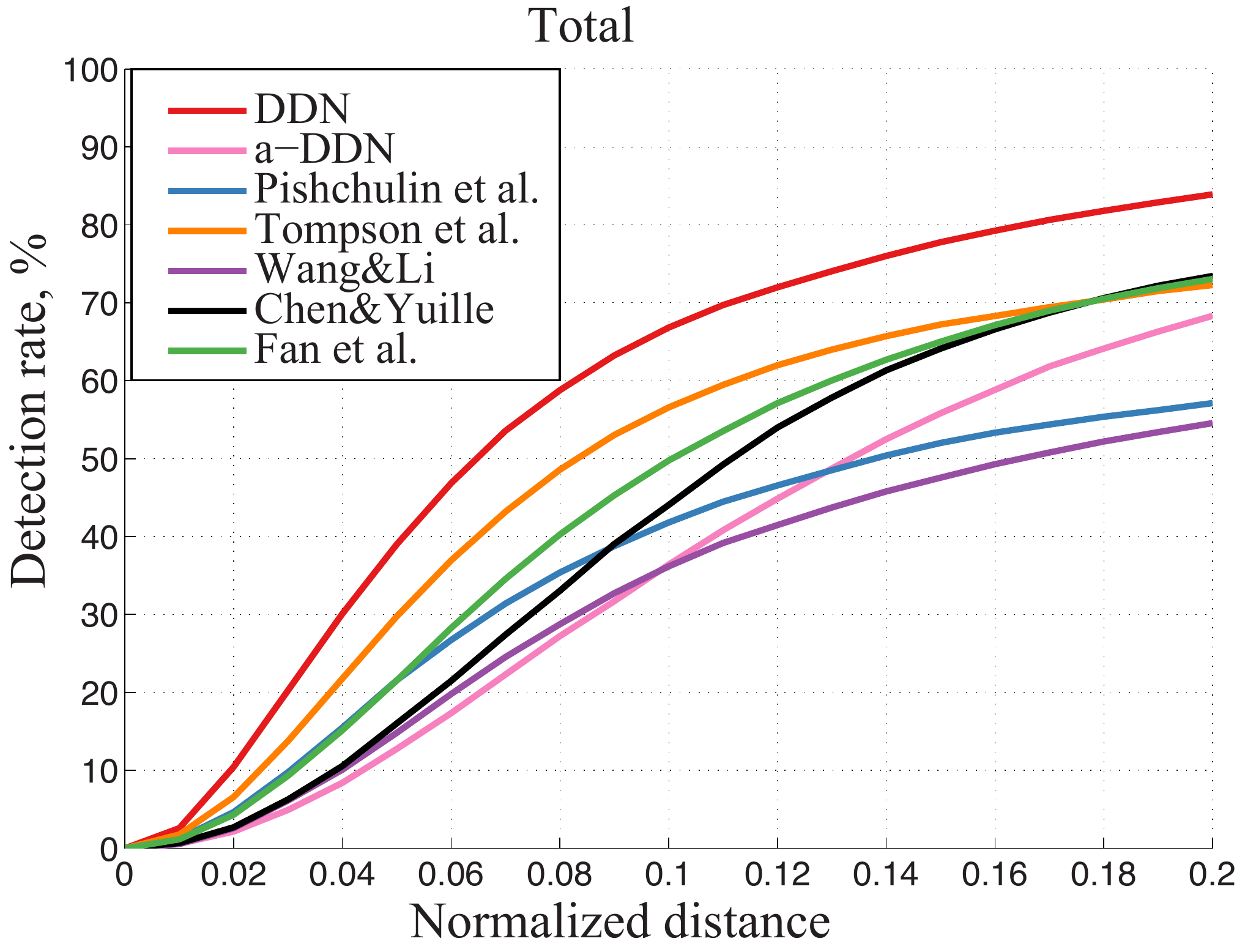}
\caption{PCK comparisons on LSP dataset with state-of-the-art methods. The horizontal axis is the normalized distance ($\alpha$) with respect to the longer dimension of the body bounding box. The vertical axis is the proportion of images in the dataset. The proposed DDN framework (red curve) outperforms previous methods by a significant margin over all $\alpha \in [0, 0.2]$, which shows the utility of an end-to-end CNN framework that incorporates geometric constraints.}
\label{fig:bodyres}
\end{figure}

\begin{table}[b]
\begin{center}
\setlength\tabcolsep{3.0pt}
\begin{tabular}{ccccccccc}
\hline
\cline{1-9}
Method & Head & Shoulder & Elbow & Wrist & Hip & Knee & Ankle & Mean \\
\hline
\cline{1-9}
Wang \& Li~\cite{fwang2013} & 84.7 & 57.1 & 43.7 & 36.7 & 56.7 & 52.4 & 50.8 & 54.6\\
Pishchulin \etal~\cite{pishchulin2013} & 87.2 & 56.7 & 46.7 & 38.0 & 61.0 & 57.5 & 52.7 & 57.1\\
Tompson \etal~\cite{tompson2014} & 90.6 & 79.2 & 67.9 & 63.4 & 69.5 & 71.0 & 64.2 & 72.0\\
Chen \& Yuille~\cite{chen2014} & 91.8 & 78.2 & 71.8 & 65.5 & 73.3 & 70.2 & 63.4 & 73.4\\
Fan \etal~\cite{fan2015} & \textbf{92.4} & 75.2 & 65.3 & 64.0 & 75.7 & 68.3 & 70.4 & 73.0\\
a-DDN (Ours) & 82.3 & 82.5 & 62.3 & 41.2 & 55.3 & 77.6 & 77.1 & 68.3\\ 
DDN (Ours) & 87.2 & \textbf{88.2} & \textbf{82.4} & \textbf{76.3} & \textbf{91.4} & \textbf{85.8} & \textbf{78.7} & \textbf{84.3} \\
\hline
\cline{1-9}
\end{tabular}
\end{center}
\caption{Comparison of PCK (\%) score at the level of 0.2 on the LSP dataset.}
\label{tb:bodynumber}
\end{table}

Figure \ref{fig:bodyres} compares our method with baselines in terms of PCK scores. In particular, the left two sub-figures show the individual performance for $2$ landmarks (Elbow and Wrist), while the right sub-figure contains the overall performance averaged on all $14$ joints. Detailed plots for all parts are provided in additional material. Among the baselines, the classical DPM-based methods~\cite{fwang2013,pishchulin2013} achieve the worst performance due to weaker low-level features. The CNN-based methods of \cite{tompson2014,chen2014,fan2015} improve over those by a large margin. However, the proposed DDN achieves a significant further improvement. PCK numbers for all the landmarks at $\alpha = 0.2$ are listed in Table \ref{tb:bodynumber}, where DDN performs better across almost all the articulated joints. The mean accuracy over all landmarks is $10.9\%$ better than the best reported result of \cite{chen2014}. 

We also report numbers on the version of DDN trained with affine transformations. It is observed that the improvement in accuracy from using TPS warps as opposed to affine transformations is significantly larger for human body parts than facial landmarks. This reflects the greater non-rigidity inherent in the human body pose estimation problem, which makes our improvement over previous CNN-based methods remarkable. 

\begin{table}[t]
\begin{center}
\setlength\tabcolsep{1.5pt}
\begin{tabular}{ccccccccccccccccc}
\hline
\cline{1-17}
$\alpha$ & Methods & Ba & Be & By & Bt & Cn & Fo & Le & Ll & Lw & Na & Re & Rl & Rw & Ta & Th\\
\hline
\cline{1-17}

\multirow{2}{*}{0.02} & \cite{ning2016} & 9.4 & 12.7 & 8.2 & 9.8 & 12.2 & 13.2 & 11.3 & \textbf{7.8} & 6.7 & 11.5 & 12.5 & 7.3 & 6.2 & 8.2 & 11.8\\

& Ours & \textbf{18.8} & \textbf{12.8} & \textbf{14.2} & \textbf{15.9} & \textbf{15.9} & \textbf{16.2} & \textbf{20.3} & 7.1 & \textbf{8.3} & \textbf{13.8} & \textbf{19.7} & \textbf{7.8} & \textbf{9.6} & \textbf{9.6} & \textbf{18.3}\\

\hline

\multirow{2}{*}{0.05} & \cite{ning2016} & 46.8 & \textbf{62.5} & 40.7 & 45.1 & 59.8 & \textbf{63.7} & 66.3 & \textbf{33.7} & 31.7 & \textbf{54.3} & 63.8 & \textbf{36.2} & 33.3 & 39.6 & 56.9\\

& Ours & \textbf{66.4} & 49.2 & \textbf{56.4} & \textbf{60.4} & \textbf{61.0} & 60.0 & \textbf{66.9} & 32.3 & \textbf{35.8} & 53.1 & \textbf{66.3} & 35.0 & \textbf{37.1} & \textbf{40.9} & \textbf{65.9}\\

\hline

\multirow{2}{*}{0.08} & \cite{ning2016} & 74.8 & \textbf{89.1} & 70.3 & 74.2 & \textbf{87.7} & \textbf{91.0} & \textbf{91.0} & 56.6 & 56.7 & \textbf{82.9} & 88.4 & 56.4 & 58.6 & 65.0 & 87.2\\

& Ours & \textbf{88.3} & 73.1 & \textbf{83.5} & \textbf{85.7} & 85.0 & 84.7 & 88.3 & \textbf{57.5} & \textbf{58.9} & 77.1 & \textbf{88.7} & \textbf{62.1} & \textbf{59.1} & \textbf{66.6} & \textbf{87.4}\\

\hline

\multirow{2}{*}{0.10} & \cite{ning2016} & 85.6 & \textbf{94.9} & 81.9 & 84.5 & \textbf{94.8} & \textbf{96.0} & \textbf{95.7} & 64.6 & 67.8 & \textbf{90.7} & 93.8 & 64.9 & \textbf{69.3} & 74.7 & \textbf{94.5}\\

& Ours & \textbf{94.0} & 82.5 & \textbf{92.2} & \textbf{93.0} & 92.2 & 91.5 & 93.3 & \textbf{69.7} & \textbf{68.1} & 86.0 & 93.8 & \textbf{74.2} & 68.9 & \textbf{77.4} & 93.4\\

\hline
\cline{1-17}

\end{tabular}
\end{center}
\caption{Comparison of PCK (\%) score on CUB200-2011. Landmark labels are abbreviated (\eg, ``Ba" denotes ``Back").}
\label{tb:birdnumber}
\end{table}

These results highlight some of the previously discussed advantages of DDN over prior CNN-based frameworks. DDN incorporates geometric structure directly into the network, which makes shape prediction during training and testing end-to-end, while also regularizing the learning. Thus, DDN can learn highly non-linear mappings, which is non-trivial with hand-designed graphical models. Further, we hypothesize that extra modules such as graphical model inference of joint neighborhoods incur additional error. The second row of Fig.~\ref{fig:res} shows several qualitative results generated by DDN, which handles a wide range of body poses with good accuracy. The challenging cases within the red box in the second row of Fig.~\ref{fig:res} show that our method degrades when the body parts are highly occluded or folded.

\subsection{Bird Part Localization}

\begin{figure}[t]
\centering
\includegraphics[width=0.95\textwidth]{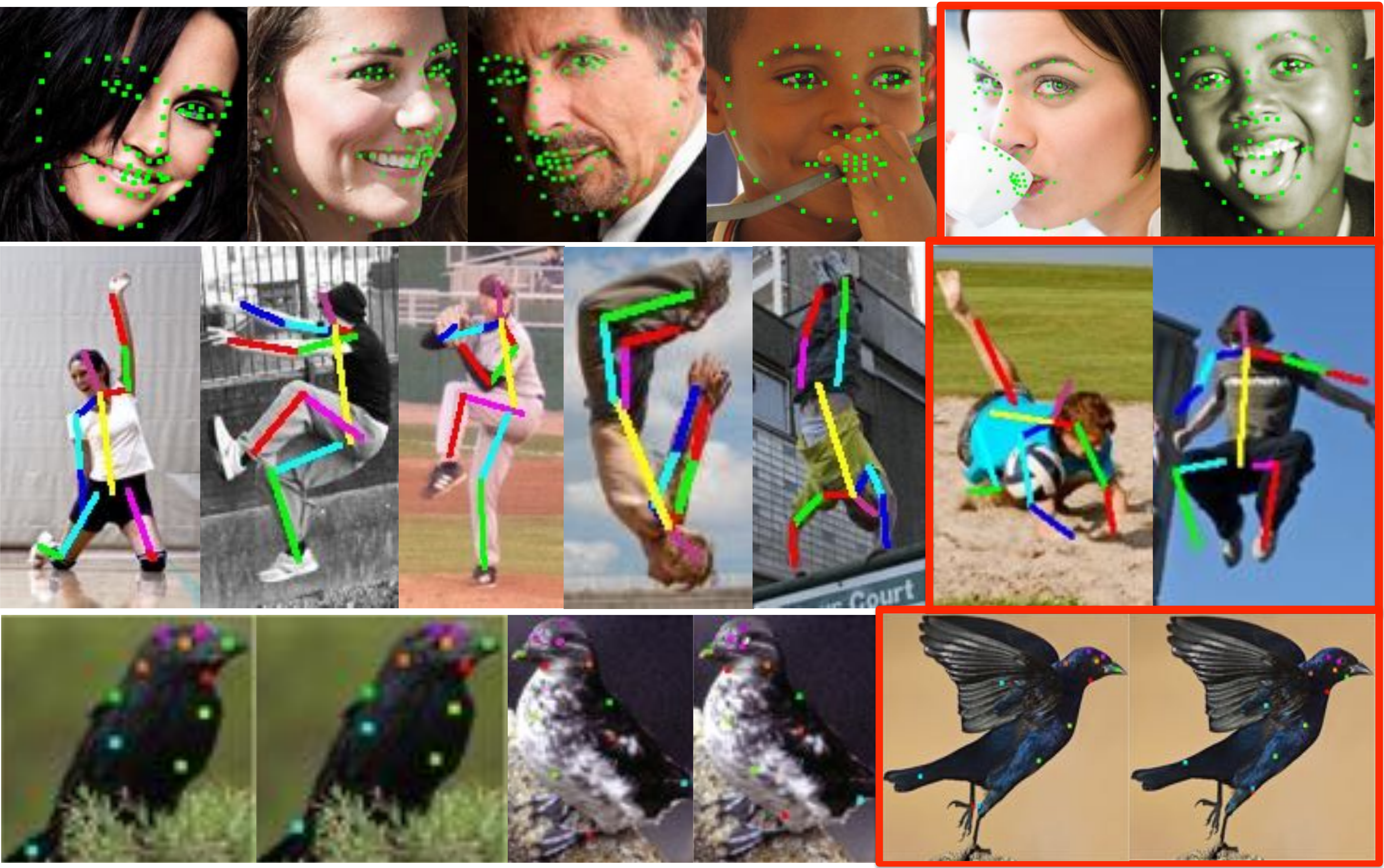}
\caption{Qualitative results of the proposed method for facial landmark localization, human body pose estimation and bird part localization. For bird landmark localization in the last row, the left image in each pair is the ground truth and the right image is our prediction. In each case, we observe that the proposed DDN achieves qualitatively good landmark localization, even with significant non-rigid deformations. Examples in the red boxes show challenging cases, for instance, occlusions around the mouth in face images, limbs in body images or wings in bird images.}
\label{fig:res}
\end{figure}

We now evaluate DDN on the well-known CUB200-2011~\cite{welinder2010} dataset for bird part localization. The dataset contains 11,788 images of 200 bird species. Each image is annotated with a bounding box and 15 key points. We adopt the standard dataset split, where 5,994 images are used for training and the remaining 5,794 for testing. CUB200-2011 was originally designed for the classification task of fine-grained recognition in the wild, this, it contains very challenging pose variations and severe occlusions. Compared to facial landmark and human body joint, another difficulty is the non-discriminative texture for many bird parts. For instance, the last row of Fig.~\ref{fig:res} shows examples where part definitions such as wings or tail might be ambiguous even for humans. The abbreviation in Table.~\ref{tb:birdnumber}

For comparison, we choose the recent work from Zhang \etal~\cite{ning2016} as the baseline. We report the PCK numbers at $\alpha = 0.02, 0.05, 0.08, 0.10$  for each of the $15$ landmarks in Table \ref{tb:birdnumber}. By converting the landmark labels to a dense segmentation mask, Zhang \etal~exploit fully convolutional networks~\cite{long2015} for landmark localization. Instead, our DDN directly regresses from the VGG features to the locations of the sparse landmarks, which incurs significantly less computational cost. In addition, Zhang \etal~predict each landmark independently without the consideration of geometric relations among landmarks, which are naturally encoded in our SBN. Therefore, our method achieves highly competitive and sometimes better performance than the state-of-the-art, at a significantly lower expense.

\section{Conclusion}

We propose a cascaded network called Deep Deformation Network (DDN) for object landmark localization. We argue that incorporating geometric constraints in the CNN framework is a better approach than directly regressing from feature maps to landmark locations. This hypothesis is realized by designing a two-stage cascade. The first stage, called the Shape Basis Network, initializes the shape as constrained to lie within a low-rank manifold. This allows a fast initialization that can account for large out-of-plane rotations, while regularizing the estimation. The second stage, called the Point Transformer Network, estimates local deformation in the form of non-rigid thin-plate spline warps. The DDN framework is trainable end-to-end, which combines the power of CNN feature representations with learned geometric transformations. In contrast to prior approaches, DDN avoids complex initializations, large cascades with several CNNs, hand-crafted features or pre-specified part connectivities of DPMs. Our DDN framework consistently achieves state-of-the-art results on three separate tasks, \ie face landmark localization, human pose estimation and bird part localization, which shows the generality of the proposed method.

\clearpage
\bibliographystyle{splncs}
\footnotesize{
\bibliography{eccvbib}
}
\end{document}